\documentclass[a4paper,fleqn]{cas-sc}

\shorttitle{}
\shortauthors{Bi et al.}

\ExplSyntaxOn
\cs_set:Npn \__first_footerline:
{
  \group_begin:
  \small\sffamily
  \__short_authors: :~
  {\rmfamily\itshape arXiv~preprint}
  \group_end:
}
\ExplSyntaxOff

\usepackage[numbers,sort&compress]{natbib}
\let\cite\citep
\usepackage{amsmath,amssymb,amsfonts}
\usepackage{algorithmic}
\usepackage{algorithm}
\usepackage{graphicx}
\usepackage{textcomp}
\usepackage{multirow}
\usepackage{booktabs}
\usepackage{xurl} 
\hypersetup{hidelinks}
\usepackage{siunitx}  
\usepackage{rotating}

\usepackage{soul}
\usepackage{etoolbox}
\usepackage{lastpage}
\AtBeginDocument{}
\usepackage{mdframed}
\newif\ifshowrevisions
\showrevisionsfalse
\sethlcolor{yellow!35}
\DeclareRobustCommand{\rev}[1]{\ifshowrevisions\hl{#1}\else#1\fi}
\soulregister\cite7
\soulregister\citep7
\soulregister\citet7
\soulregister\ref7
\soulregister\url7
\newif\ifrevisionbibopen
\newcommand{\closerevisionbib}{%
  \ifrevisionbibopen
    \global\revisionbibopenfalse
    \end{mdframed}%
  \fi}
\newcommand{\revisionbibitem}[2][]{%
  \closerevisionbib
  \originalbibitem[#1]{#2}%
  \ifstrequal{#2}{bi2025mamba}{%
    \ifshowrevisions
      \begin{mdframed}[backgroundcolor=yellow!35,linewidth=0pt,
        innerleftmargin=0pt,innerrightmargin=0pt,
        innertopmargin=2pt,innerbottommargin=2pt,
        skipabove=0pt,skipbelow=0pt]
      \global\revisionbibopentrue
    \fi}{}%
  \ignorespaces}
\AtBeginEnvironment{thebibliography}{%
  \let\originalbibitem\bibitem
  \let\bibitem\revisionbibitem}
\AtEndEnvironment{thebibliography}{\closerevisionbib}

\newcommand{\best}[1]{{\color{red}\bfseries #1}}
\newcommand{\second}[1]{{\color{blue}\bfseries #1}}

\begin{document}
\renewcommand{\printorcid}{}
\title[mode=title]{Data-Augmented State-Space Modelling for Fertiliser-Aware Decision Support in Precision Agricultural Automation}

\author[1]{Gaoshan Bi}
\author[1]{Tong Liu}
\author[2]{Yu Zhang}
\author[3]{Qing Xue}
\author[3]{Yang Li}
\author[1]{Po Yang}
\cormark[1]
\ead{po.yang@sheffield.ac.uk}

\affiliation[1]{
  organization={School of Computer Science, University of Sheffield},
  city={Sheffield},
  country={UK}
}

\affiliation[2]{
  organization={College of Computer Science and Technology, Zhejiang University of Technology},
  city={Hangzhou},
  country={China}
}

\affiliation[3]{
  organization={Research and Development, Mutus Tech Ltd},
  city={Liverpool},
  country={UK}
}


\cortext[1]{Corresponding author.}

\begin{abstract}
Accurate prediction of fertiliser-induced nitrous oxide (N$_2$O) emissions is crucial for optimising sustainable fertiliser strategies that reduce greenhouse gas (GHG) emissions while maintaining crop yields.
However, most existing studies struggle to effectively capture long-range temporal dependencies with limited agricultural data, and lack integration with agricultural management systems that combine N$_2$O prediction with data management and precision fertilisation.
In this work, we present DA-Mamba, a data augmentation-enhanced state space framework for single- and multi-step prediction of fertiliser-induced N$_2$O emissions. The framework captures long-range temporal dynamics and high-order interactions among multivariate agro-environmental variables.
On public fertiliser-induced N$_2$O datasets, DA-Mamba achieves state-of-the-art results, improving single-step prediction accuracy by up to 12.0\% (RMSE) and 25.8\% (MAE) over six baselines, and multi-step forecasting by 2.0\% (RMSE) and 4.8\% (MAE) over three baselines.
We further deploy DA-Mamba as a containerised microservice within ParallelFarm, a precision fertilisation and farm management system, which ingests weather, soil, and management data in real-time to generate optimised fertiliser plans and field-level N$_2$O forecasts.
Predictions are continuously refined through ongoing on-farm trials and automated retraining in a closed-loop workflow, with chamber-based measurements from field sites in the United Kingdom supporting external validation and dataset expansion.
\end{abstract}

\begin{keywords}
Fertiliser-induced N$_2$O prediction \sep sustainable agricultural system \sep state space model \sep Mamba
\end{keywords}

\maketitle
\hypersetup{pdfauthor={Gaoshan Bi; Tong Liu; Yu Zhang; Qing Xue; Yang Li; Po Yang}}

\section{Introduction}
\label{sec:introduction}
Nitrous oxide (N$_2$O) is a potent greenhouse gas (GHG) with a global warming potential (GWP) 265 times greater than that of carbon dioxide (CO$_2$)~\cite{xiao2024spatiotemporal}.
Atmospheric N$_2$O concentrations are rising at a rate of 0.98 ppb yr$^{-1}$, which is 44\% higher than in 2005~\cite{saha2021machine}, making it a significant driver of climate change.
Existing research shows that nitrogen fertiliser use in agriculture contributes roughly 50\% of global anthropogenic N$_2$O emissions~\cite{shcherbak2014global}, a share that is rising rapidly.
Excessive fertiliser application aimed at boosting crop yields further accelerates this trend~\cite{zhang2021spatio}.
Consequently, it is essential to reduce N$_2$O emissions while sustaining yields by implementing optimised fertilisation strategies~\cite{liu2024co}.
Achieving this objective necessitates precise measurement of fertiliser-induced N$_2$O emissions, thereby enabling informed adoption of sustainable fertilisation practices in line with the paradigms of Agriculture 4.0~\cite{liu2020industry} and intelligent precision agriculture~\cite{shu2021guest}.
In this context, chamber-based techniques have been widely adopted as a standard approach for N$_2$O flux monitoring due to their high measurement accuracy. Recent advances further integrate chamber systems with wireless sensor networks and agricultural IoT platforms, enabling near real-time data acquisition and supporting the digital transformation of modern agricultural systems~\cite{chadwick2014optimizing, saranya2023comparative, patel2023review}.
However, each chamber captures N$_2$O fluxes over only a limited spatial footprint, and large-scale deployment remains impractical due to high installation and maintenance costs~\cite{knappe2021assessing}. Moreover, chamber-based monitoring is inherently retrospective, providing post-emission measurements without the predictive capability required for proactive fertilisation management. These limitations underscore the need for scalable, cost-effective, and predictive approaches to complement chamber-based monitoring frameworks.
To overcome the limitations of chamber-based monitoring, process-based biogeochemical models such as DNDC have been used to simulate fertiliser-induced GHG emissions by explicitly modelling soil–plant–atmosphere processes \cite{yin2020carbon}.
Nevertheless, their dependence on predefined process representations and expert-driven parameterisation constrains flexibility and generalisation across diverse agricultural settings.

With advances in data-driven methods, machine learning (ML) have become increasingly popular for smart agriculture~\cite{akkem2023smart}.
For instance, aerial visual perception frameworks have been successfully deployed for crop disease monitoring~\cite{su2020aerial}, and biometric identification systems have facilitated individuality-based precision aquaculture~\cite{schraml2020towards}.
Inspired by these successes in processing complex farming data, ML is also being applied to modelling N$_2$O fluxes in agricultural systems, particularly during nitrogen fertiliser application periods~\cite{petersen2023higher}.
Ensemble-based algorithms such as random forest (RF)~\cite{saha2021machine} and extreme gradient boosting (XGBoost)~\cite{wang2024sources}have been widely applied to relate N$_2$O emission measurements to environmental and management variables, including meteorological conditions, soil characteristics, and fertiliser usage~\cite{liu2024co, vemuri2024developing}.
To better capture temporal dependencies and emission spikes, recurrent and attention-based deep learning architectures, such as long short-term memory (LSTM) network~\cite{hamrani2020machine} and Transformer-based model~\cite {lin2024agriculture}, have demonstrated improved accuracy over RF by modelling dynamic emission patterns.
In addition, knowledge-guided machine learning (KGML) has emerged, integrating domain-specific agricultural insights into gated recurrent unit (GRU) networks~\cite{liu2022kgml}.
By pre-training on process-based synthetic datasets and fine-tuning with limited field observations, KGML achieved notable improvements in fertiliser-induced N$_2$O emission prediction.
More recently, physically informed neural architectures, such as physics-constrained multilayer perceptrons (MLPs), have been developed to strengthen generalisation performance across diverse agricultural settings~\cite{vemuri2024developing}.

The aforementioned studies have demonstrated the potential of learning-based methods to model observations from experimental fields and predict N$_2$O emissions resulting from fertilisation.
However, they typically focus on modelling N$_2$O emissions under various fertiliser strategies and farming factors without simultaneously considering the corresponding yields.
Consequently, these findings alone are insufficient for devising optimal fertilisation strategies in real-world agricultural practices that simultaneously sustain crop yields and reduce emissions.
Moreover, the majority of existing studies rely on either recurrent neural networks (RNNs) or Transformer-based architectures.
In the complex and heterogeneous conditions of real-world agricultural environments and fertilisation practices, RNNs frequently struggle to capture the long-range temporal dynamics of N$_2$O fluxes~\cite{zhao2020rnn, liu2020fast, liu2023adaptive}.
While Transformers are capable of modelling such dependencies, they incur quadratic time–memory complexity and typically require large-scale training corpora, which can be impractical for agricultural applications where labelled data is scarce~\cite{liuitransformer, lu2024application}.
These limitations highlight the need for alternative architectures that combine the capacity to model long-range dependencies with computational efficiency and robustness under data-limited conditions~\cite{gu2020hippo, gu2021efficiently}.
Furthermore, there is a notable lack of integrated sustainable agricultural management systems that effectively combine N$_2$O flux prediction frameworks with agricultural data management and precision fertilisation prediction, both of which are essential for the advancement of real-world agricultural management practices~\cite{dhanaraju2022smart, khan2021current}.

Recent state-space architectures, such as Mamba~\cite{gu2023mamba, mamba2} and its variants~\cite{wang2025mamba, wang2025effective}, are able to effectively model long series and are well-suited to capturing complex dependencies in multivariate time series.
However, these architectures have not yet been applied to modelling fertiliser-induced nitrous oxide emissions.
Based on these studies, we propose a data augmentation-enhanced state space model (DA-Mamba) that predicts N$_2$O emissions by leveraging the environmental data and precision fertilisation plans.
Specifically, the DA-Mamba framework consists of two main components: (i) a data augmentation module, which enables robust generalisation from limited agricultural datasets, and (ii) a bi-directional Mamba regression module, which simultaneously captures complex dependencies among multivariate agro-environmental variables and the temporal dynamics of N$_2$O emissions.
Additionally, we demonstrate real-world integration by deploying DA-Mamba as a containerised microservice within an existing sustainable farming management system~\cite{bi2024parallelfarm, zhang2023sustainable, zhang2022spatio}.
Finally, we validate the framework through ongoing field trials in the UK, where chamber-based N$_2$O flux measurements provide external verification and facilitate dataset expansion.
The core single-step prediction algorithm was initially reported in our previous conference version~\cite{bi2025mamba}, and the present work substantially extends it.
Specifically, we have augmented the study with multi-step prediction tasks, comprehensive parameter evaluations and visual analyses, and a practical microservice integration within the ParallelFarm system.
The main contributions of this work are as follows:
\begin{itemize}
\item We present a unified data integration approach that combines climate, soil, and management data to support robust modelling of agricultural N$_2$O emissions, compiling multi-source datasets for model development and validation.
\item We introduce enhanced state space models with data augmentation for single- and multi-step prediction of post-fertilisation N$_2$O emissions. These models effectively capture temporal dependencies and multivariate interactions, and include a comprehensive parameter sensitivity analysis, achieving state-of-the-art performance compared with multiple benchmark models.
\item We deploy the models within the ParallelFarm, a precision fertilisation 
and farm management system (accessible at \url{https://app.parallelfarm.co.uk}).
This enables spatially generalisable N$_2$O predictions and is applied in on-farm trials in the UK, where continuous flux measurements support dynamic model fine-tuning and real-world validation.
\end{itemize}

\section{Data Collection and Preparation}
\subsection{Experiment and Data Description}
We have identified the agricultural environmental variables and farming records that are necessary for the prediction of N$_2$O emissions after fertilisation, based on the previous works\cite{hamrani2020machine, saha2021machine, liu2022kgml}.
As shown in Table~\ref{variables_description}, the model necessitates seven climate variables, six soil-related properties, and two crop practice records.
An additional advantage of selecting these variables is that our system can readily obtain them from the ERA5-Land weather dataset\cite{hersbach2020era5} and SoilGrids250m soil dataset\cite{hengl2017soilgrids250m}, thereby enhancing its scalability and providing a foundation for large-scale sustainable agricultural services.
After that, we proceeded to extract the data from previous research for model training and preliminary validation.
In particular, the N$_2$O dataset was originally collected using chambers in six controlled-environment experimental fields at the University of Minnesota\cite{liu2022kgml}.
Six chambers—each with a soil surface area of 2 m$^2$ and a column depth of 1.1 m, under continuous corn cultivation—were utilised to monitor hourly N$_2$ fluxes (mg N m$^{-2}$ h$^{-1}$) in response to varying climate conditions, soil properties, and nitrogen fertiliser practices during 2016, 2017, and 2018. The collection period spanned from 1 April to 31 July each year, covering 122 days.
This time frame was chosen based on the typical early-May initiation of corn planting and the pivotal role of nitrogen fertiliser in modulating N$_2$O flux dynamics.


\begin{table}[pos=!t]
    \centering
    \caption{Environmental Variables and Farming Data.}
    \label{variables_description}
    \resizebox{0.7\textwidth}{!}{
        \begin{tabular}{ll}
            \toprule
            \textbf{Data variables} & \textbf{Description} \\
            \midrule
            
            \multirow{7}{*}{Climate variables} 
                & Surface downward shortwave radiation (W m$^2$) \\
                & Maximum air temperature ($^\circ$C) \\ 
                & Difference of max and min air temperature ($^\circ$C) \\ 
                & Maximum humidity (fraction) \\
                & Difference of max and min humidity (fraction) \\ 
                & Wind speed (m s$^{-1}$) \\ 
                & Precipitation (mm day$^{-1}$) \\
            \cmidrule(lr){1-2}
            
            \multirow{6}{*}{Soil properties} 
                & Soil averaged bulk density (Mg m$^{-3}$) \\ 
                & Soil averaged sand content (g kg$^{-1}$) \\
                & Soil averaged silt content (g kg$^{-1}$) \\ 
                & Soil averaged pH (unitless) \\
                & Cation exchange capacity of soil (cmol$^-$ kg$^{-1}$) \\ 
                & Soil averaged organic carbon (g C kg$^{-1}$) \\ 
            \cmidrule(lr){1-2}
            
            \multirow{2}{*}{Crop practice records} 
                & Amount and fertiliser date (g N m$^2$) \\ 
                & Plant day of the year \\
            \bottomrule
        \end{tabular}
    }
\end{table}

Additionally, based on the experiment definition, we are carrying out the fertiliser N$_2$O emission trials on Velcourt farms located in Lincolnshire ($52.667^\circ$ N, $-0.522^\circ$ W) and Cambridgeshire ($52.010^\circ$ N, $0.051^\circ$ W) in the UK.
As shown in Fig.~\ref{SystemFramework}, chambers were installed in the test fields to collect N$_2$O data both before and after fertiliser application.
These measurements will be compared with data simulated by the system, and the framework is intended to be further refined using the real-world data collected.
A major challenge in this approach is the time-consuming nature of collecting post-fertilisation data using chambers in test fields, as each tillage cycle yields only a limited dataset.
Therefore, in line with the experimental setup, the necessary data were extracted from public databases to support both the training and validation of the proposed flux prediction model.
This strategy also equips the model with few-shot capabilities in predicting post-fertilisation GHG fluxes, requiring only minimal fine-tuning with a few future samples---or even achieving zero-shot performance.

\subsection{Problem Formulation}
In the N$_2$O flux prediction task involving nitrogen fertiliser, the proposed DA-Mamba framework receives as input a historical daily time series of climate, soil, and crop management data, denoted as
\begin{equation}
X^{\text{input}} = [x_1, x_2, \dots, x_L] \in \mathbb{R}^{L \times P},
\end{equation}
where $L$ is the input sequence length, $x_n = [v_1, v_2, \dots, v_P]$ represents the $P$ input variates at day $n$. The corresponding output sequence of daily N$_2$O fluxes is
\begin{equation}
Y^{\text{output}} = [y_1, y_2, \dots, y_L]^\top \in \mathbb{R}^{L \times 1}.
\end{equation}

\textit{Single-step sequence-to-sequence regression.}  
Given the input $X^{\text{input}}$, the objective is to learn a mapping
\begin{equation}
\hat{Y}^{\text{output}} = f(X^{\text{input}};\,\theta),
\end{equation}
where $f(\cdot)$ is the DA-Mamba model parameterised by $\theta$, and $\hat{Y}^{\text{output}}$ denotes the predicted flux sequence aligned with the input window.

\textit{Multi-step forecasting.}  
Beyond aligned sequence regression, the framework can also be extended to forecast $N$ future steps beyond the observed window:
\begin{equation}
Y^{\text{future}} = [y_{L+1}, y_{L+2}, \dots, y_{L+N}]^\top \in \mathbb{R}^{N \times 1},
\end{equation}
with predictions
\begin{equation}
[\hat{y}_{L+1}, \hat{y}_{L+2}, \dots, \hat{y}_{L+N}]^\top = f_{\text{ms}}(X^{\text{input}};\,\theta).
\end{equation}

\textit{Optimisation objective.}  
During training, DA-Mamba minimises the mean squared error (MSE) between the predicted and ground-truth fluxes. For sequence-to-sequence regression:
\begin{equation}
\mathcal{L}_{\text{seq}}(\theta) = \frac{1}{L}\sum_{i=1}^{L}(y_i - \hat{y}_i)^2,
\end{equation}
and for multi-step forecasting:
\begin{equation}
\mathcal{L}_{\text{ms}}(\theta) = \frac{1}{N}\sum_{i=1}^{N}(y_{L+i} - \hat{y}_{L+i})^2.
\end{equation}

At evaluation, we report RMSE, MAE, and R$^2$ as defined in Section~\ref{sec:experiments} to comprehensively assess predictive performance.

\section{Proposed Method}
\subsection{Overview of Proposed Framework}
\begin{figure}[pos=!htbp]
\centering
\includegraphics[width=1.0\linewidth]{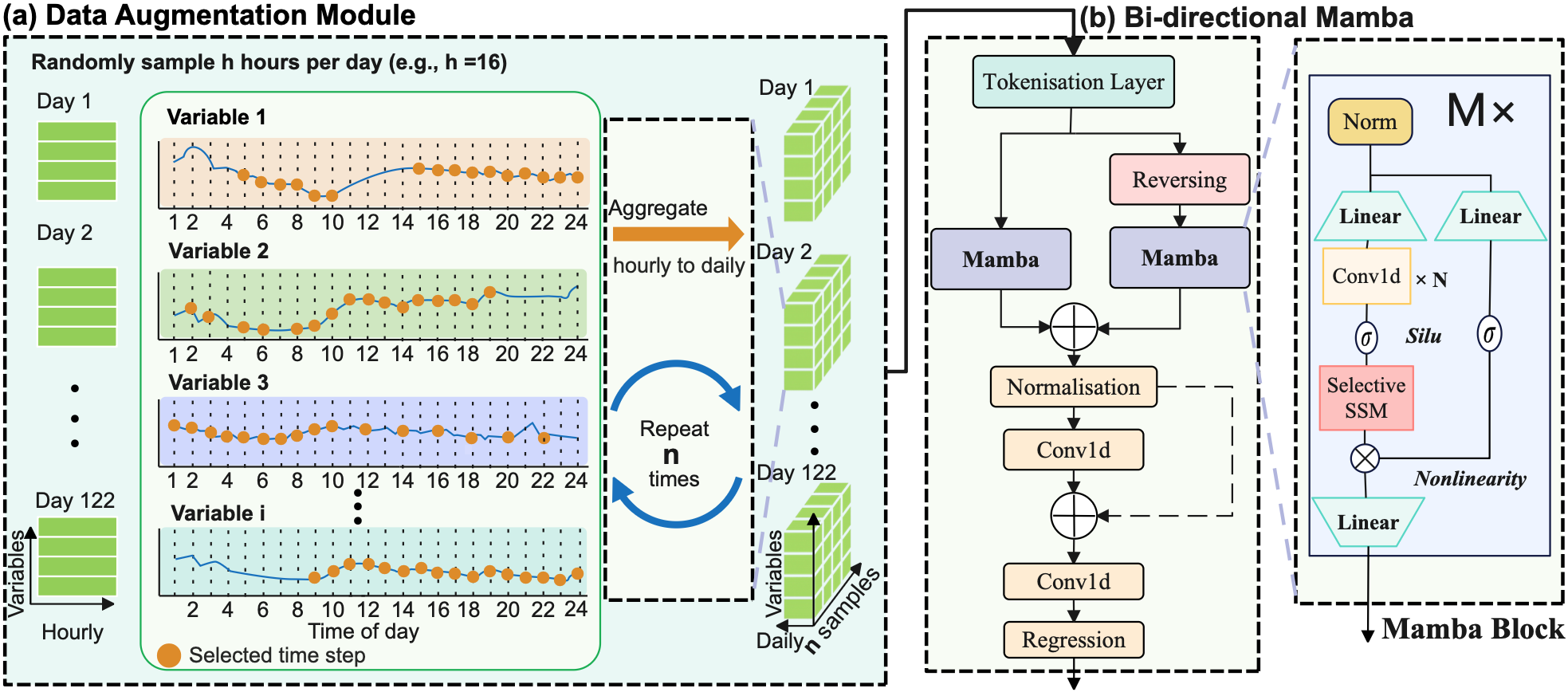}
\caption{The data augmentation-enhanced Mamba framework.}
\label{Framework}
\end{figure}

As illustrated in Fig.~\ref{Framework}, the framework begins by feeding time-series raw data from agricultural practices into a dedicated data augmentation module.
This module tackles the challenge of limited agricultural datasets by applying augmentation strategies that mitigate overfitting and improve the model’s ability to generalise.
The augmented data is then normalised and tokenised before entering the encoding module.
At the heart of the encoder are the Mamba block and a one-dimensional convolutional layer, which are responsible for capturing inter-variable relationships and temporal dependencies, respectively.
After passing through $M$ encoding layers, the output is transformed by a linear projection layer to generate the daily N$_2$O predictions.

\subsection{Data Augmentation Module}
To mitigate overfitting and improve the generalisation of our model, which was trained on a relatively small dataset of farming practices, we have incorporated a data augmentation module inspired by the work of Liu et al.~\cite{liu2022kgml} into the proposed framework.
This module enlarges the effective sample size by a factor of $n$.
In our setting, climate variables are recorded hourly and aggregated into daily values. From each 24-hour period, $h$ valid hourly observations are randomly selected and averaged to produce a representative daily measurement.
To preserve diurnal patterns while introducing minimal variation, $h$ is set between 12 and 24 hours.
The N$_2$O observation ratio is computed as $\frac{24 - (\text{number of missing hours})}{24}$ and incorporated as a weight in the loss function, ensuring that data quality is considered during optimisation and allowing effective use of records with missing N$_2$O values.
If 16 or more hours are missing in a given day, the record is deemed unreliable, its weight is set to zero, and it is excluded from training.
The module processes both input features and prediction targets for the training set, whereas for the test set only the N$_2$O observation ratio is calculated.
This procedure transforms the original dataset into an $n$-fold expanded version, thereby improving model robustness.

\subsection{Mamba Block}
Mamba is a specialised State Space Model (SSM) that adaptively updates its state space parameters based on the input and selectively processes tokens according to the current context~\cite{gu2023mamba, mamba2}.
As a consequence of their foundation in Kalman filter theory, SSMs are well suited to the modelling of cyclical processes through latent states.
They describe the evolution of these states and their mapping to outputs using first-order differential equations, as shown in Eq.~\eqref{first_ODE}:
\begin{equation}
h_t = A h_{t-1} + B x_{t}, \quad y_t = C h_t
\label{first_ODE}
\end{equation}

The structured state space model (S4), an extension of the vanilla SSM, applies HiPPO-based initialisation to constrain the state matrix, enhancing the ability to capture long-range dependencies~\cite{gu2021efficiently}. To handle continuous sequences, a timescale parameter $\Delta$ is introduced to discretise the system, converting the continuous parameters $A$ and $B$ into their discrete forms $\overline{A}$ and $\overline{B}$. The zero-order hold (ZOH) is commonly used for this transformation:
\begin{equation}
\overline{A} = \exp(\Delta A),
\label{2}
\end{equation}
\begin{equation}
\overline{B} = (\Delta A)^{-1}\bigl(\exp(\Delta A) - I\bigr)\cdot\Delta B
\label{3}
\end{equation}

With step size $\Delta$, the discretised system becomes:
\begin{equation}
h_t = \overline{A} h_{t-1} + \overline{B} x_{t-1}, \quad y_t = C h_t
\end{equation}
where $\overline{A} \in \mathbb{R}^{N\times N}$ is the evolution parameter, and $\overline{B} \in \mathbb{R}^{N\times 1}$, $C \in \mathbb{R}^{1\times N}$ are the projection parameters. The output is then obtained via a global convolution, enabling parallel computation:
\begin{equation}
\overline{K}
= \bigl(C\overline{B},, C\overline{AB},, C\overline{AAB}, \dots, C\overline{A^{L-1}B}\bigr)
\label{K_C_B}
\end{equation}
\begin{equation}
y = x * \overline{K},
\label{y_x_k}
\end{equation}
where $L$ is the input sequence length, and $\overline{K} \in \mathbb{R}^L$ acts as the structured convolution kernel.

\subsection{Bi-directional Mamba Regression Algorithm}
\begin{algorithm}[!b]
  \caption{Bi-directional Mamba regression Algorithm}
  \label{al:Mamba_encoding}

  {\footnotesize
  \setlength{\baselineskip}{0.6\baselineskip}

  \begin{algorithmic}[1]
    \renewcommand{\algorithmicrequire}{\textbf{Input:}}
    \renewcommand{\algorithmicensure}{\textbf{Output:}}

    \REQUIRE $Batch(X^{in}) = [x_1, x_2, \dots, x_L] \in (B, L, P)$
    \ENSURE (1) $Batch(Y^{out}) = [y_1, y_2, \dots, y_L] \in (B, L, 1)$ for single-step regression
    \ENSURE (2) $Batch(Y^{future}) = [y_{L+1}, y_{L+2}, \dots, y_{L+N}] \in (B, N, 1)$ for multi-step forecasting

    \STATE $Batch(X^{in}) \gets$ Standardise($Batch(X^{in})$)
    \STATE \textbf{Linear Tokenisation Layer:}
    \STATE $Batch(X) \in (B, P, L) \gets$ Permute($Batch(X^{in}) \in (B, L, P)$)
    \STATE $X^{token} \in (B, P, D) \gets$ Tokenise($Batch(X)$)

    \STATE \textbf{for} $i$ \textbf{in} numLayers \textbf{do}:
    \STATE \quad \textbf{Bi-directional Mamba Encoding}
    \STATE \quad $Z^{forward} \in (B, P, D) \gets$ Mamba($X^{token}$)
    \STATE \quad $X^{backward} \in (B, P, D) \gets$ Flip[dims=1]($X^{token}$)
    \STATE \quad $Z^{backward} \in (B, P, D) \gets$ Mamba($X^{backward}$)
    \STATE \quad $Z' \in (B, P, D) \gets Z^{forward} + Z^{backward}$

    \STATE \quad \textbf{Residual Connection}
    \STATE \quad $Z' \in (B, P, D) \gets Z' + X^{token}$
    \STATE \quad $Z' \in (B, P, D) \gets$ Normalise($Z'$)

    \STATE \quad \textbf{1D Convolution Extracts Temporal Correlations}
    \STATE \quad $U \in (B, D, P) \gets$ Permute($Z' \in (B, P, D)$)
    \STATE \quad $U \in (B, D, P) \gets$ Convolute($U$)
    \STATE \quad $U \in (B, D, P) \gets$ ReLU($U$)
    \STATE \quad $U \in (B, P, D) \gets$ Permute($U \in (B, D, P)$)

    \STATE \quad \textbf{Residual Connection and Normalisation}
    \STATE \quad $U \in (B, P, D) \gets$ Normalise($U + Z'$)
    \STATE \textbf{end for}

    \STATE \textbf{Projection:}
    \STATE \quad \textit{Single-step regression (seq2seq):}
    \STATE \quad $Y \in (B, P, L) \gets$ Convolute($U \in (B, P, D)$)
    \STATE \quad $Y \in (B, L, P) \gets$ Permute($Y \in (B, P, L)$)
    \STATE \quad $Y^{out} \in (B, L, 1) \gets$ LinearProject($Y \in (B, L, P)$)

    \STATE \quad \textit{Multi-step forecasting:}
    \STATE \quad $Y \in (B, P, N) \gets$ Convolute($U \in (B, P, D)$)
    \STATE \quad $Y \in (B, N, P) \gets$ Permute($Y \in (B, P, N)$)
    \STATE \quad $Y^{future} \in (B, N, 1) \gets$ LinearProject($Y \in (B, N, P)$)

    \STATE \textbf{Return} $Y^{out}$ \textbf{or} $Y^{future}$
  \end{algorithmic}
  } 
\end{algorithm}
The proposed framework operates on augmented multivariate agricultural inputs, beginning with independent normalisation applied to each variable to ensure consistent scaling across heterogeneous data sources. Building on the concept of S-Mamba~\cite{wang2025mamba}, which extends the iTransformer~\cite{liuitransformer} paradigm, the model transforms the multivariate sequence into a high-dimensional representation by projecting each variable through a dedicated linear mapping. This step enables the separation and enhancement of variable-specific semantic information.
To capture complex temporal dependencies, the model applies Mamba encoding in both forward and reverse directions. The resulting bidirectional representations are aggregated through element-wise summation, integrated with residual pathways, and normalised to promote effective inter-variable information fusion. This bidirectional encoding mechanism serves as a lightweight yet powerful alternative to conventional attention-based architectures, offering improved scalability and interpretability in agricultural contexts.
Temporal patterns are subsequently extracted using a stack of one-dimensional convolutional layers, each equipped with residual shortcuts and layer normalisation to stabilise learning and facilitate deeper feature propagation. After $M$ iterations, the encoded sequence is temporally compressed through a convolutional adjustment layer, aligning the sequence length with the prediction horizon. A final linear projection maps the output to the desired time series target dimension.
This architecture is specifically tailored to capture the evolution of fertiliser-induced N$_2$O emissions, accounting for temporal correlations driven by agricultural practices, environmental conditions, and management interventions. The integration of residual mechanisms and normalisation not only supports robust convergence during training but also enhances generalisability across varied agricultural scenarios.
For a comprehensive overview of the procedural steps, please refer to \textbf{Algorithm~\ref{al:Mamba_encoding}}. 

\section{Experiments}
\label{sec:experiments}
In this section, we perform extensive experiments on the real-world fertiliser N$_2$O flux dataset to evaluate the proposed DA-Mamba.
For further details regarding the dataset, please refer to Section 2.

\subsection{Experimental Setup}
We evaluate our proposed method against six recent state-of-the-art ML benchmarks for predicting fertiliser-induced N$_2$O emissions.
These include three conventional models-random forest (RF)~\cite{saha2021machine, liu2024co}, extreme gradient boosting (XGBoost)\cite{wang2024sources}, and multi-layer perceptron (MLP)~\cite{vemuri2024developing}-which do not account for temporal dependencies, and instead focus solely on modelling the relationships between multiple input features and the target variable.
In addition, we incorporate three sequential models-gated recurrent units (GRUs)~\cite{liu2022kgml}, long short-term memory (LSTM)~\cite{hamrani2020machine}, and Encoder-only Transformer~\cite{lin2024agriculture}-to support sequence-based regression and provide further comparative insights.
To assess the effectiveness of the data augmentation module, we further integrate it into each of the six baseline models, referring to the resulting variants with the “DA-” prefix.
To the best of our knowledge, this work represents the first attempt to apply a Mamba-based model to predict fertiliser-induced N$_2$O emissions, while also systematically exploring the combined influence of agricultural variable interactions and temporal dynamics on N$_2$O emissions.
Due to the limited scale of the N$_2$O dataset, K-fold cross-validation was adopted when partitioning the data into training and test sets.
Specifically, data collected from five chambers over the years 2016, 2017, and 2018 was used as the training set for each fold. 
Each chamber was monitored for a period of 122 days (from April 1st to July 31st), and —the data from each period was passed through the data augmentation module, resulting in a total of $n \times 5 \times 3 \times 122$ data points.
To ensure fairness in the comparative experiments, we set $n = 1000$, following the approach of Liu et al. \cite{liu2022kgml}.
The data from the remaining chamber was reserved as the test set, comprising $3 \times 122$ data points in total.
In addition, we utilise climate and soil data from the 30 days leading up to fertiliser application, inclusive of the fertilisation day, as the input sequence for multi-step prediction tasks.
The six measurement chambers shared the same fertilisation dates within each respective year, which were 4 May 2016, 3 May 2017, 8 May 2018.
The objective was to predict N$_2$O emissions for the following $N$ days, where $N \in {10, 20, 30, 40, 50}$.
All experiments in this study were conducted on an Ubuntu-based system equipped with an AMD Ryzen 7 7700X processor and an NVIDIA RTX 4060 GPU.

\subsection{Experimental Metrics}
To evaluate the predictive performance of our proposed method on the real-world N$_2$O flux dataset, we adopt three key metrics: root mean squared error (RMSE), as defined in Eq. \eqref{eqRMSE}; mean absolute error (MAE), in Eq. \eqref{eqMAE}; and the coefficient of determination (R²), as presented in Eq. \eqref{eqCoefficient}.
These metrics offer a comprehensive evaluation across different aspects of model performance, particularly reflecting the model’s capacity to capture the dynamic behaviour of N$_2$O fluxes.

\begin{equation}
\text{RMSE} = \sqrt{\frac{1}{n} \sum_{i=1}^n (y_i - \hat{y}_i)^2}\label{eqRMSE}
\end{equation}

\begin{equation}
\text{MAE} = \frac{1}{n} \sum_{i=1}^n |y_i - \hat{y}_i|\label{eqMAE}
\end{equation}

\begin{equation}
\text{R$^2$} = 1 - \frac{\sum_{i=1}^n (y_i - \hat{y}_i)^2}{\sum_{i=1}^n (y_i - \bar{y})^2}\label{eqCoefficient}
\end{equation}

\subsection{Single-Step Prediction}
\begin{figure}[pos=!b]
\centering
\includegraphics[width=0.6\linewidth]{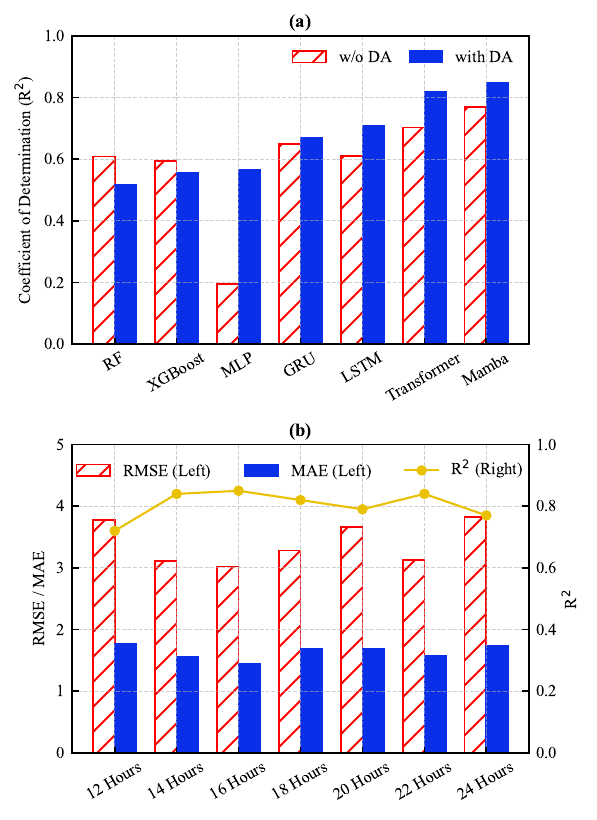}
\caption{(a) The results of k-fold cross-validation of R$^2$. 
(b) The parameter sensitivity analysis of data augmentation module.}
\label{k-fold_result}
\end{figure}

A detailed comparison of the proposed DA-Mamba model with six benchmark methods, along with their variants incorporating the data augmentation (DA) module, is presented in Table~\ref{results}.
The models were evaluated across six experimental chambers using two performance metrics: RMSE and MAE.
The results clearly indicate that DA-Mamba achieves the highest accuracy (highlighted in red), yielding the lowest RMSE and MAE in Chambers 2, 4, 5, and 6. It also ranks second (highlighted in blue) in Chamber 1's RMSE, as well as both RMSE and MAE in Chamber 3, demonstrating strong and consistent performance across diverse agricultural conditions.
The Transformer model with the DA module also shows competitive performance, benefiting from its attention mechanism for capturing temporal correlations. It achieves state-of-the-art results in Chamber 1 and Chamber 3's RMSE.
Moreover, in most cases, the best and second-best predictions are achieved by deep learning models enhanced with the DA module, demonstrating its effectiveness. The only exception is the second-best performance in Chamber 4, which was achieved without DA.
By contrast, traditional machine learning models such as RF and XGBoost exhibit degraded performance when combined with the DA module.
For example, in Chamber 4, the MAE of the RF model increased from 2.777 to 2.912 after applying data augmentation, suggesting that the DA module may introduce noise or instability for traditional ML models.
Figure~\ref{k-fold_result}(a) compares the average performance across six chambers, measured by R$^2$, for six baseline models and the proposed Mamba-based model, both with and without the DA module, where R$^2$ values closer to 1 indicate better alignment between predicted and actual outcomes.

\begin{table}[pos=!t]
\centering
\caption{Performance comparison of the proposed method and baseline models on N$_2$O emissions prediction using real-world agricultural data. (\best{Red} indicates the best results, while \second{Blue} represents the second-best results.) \rev{The results were originally reported in our previous conference version~\mbox{\cite{bi2025mamba}}.}}
\label{results}

\small
\setlength{\tabcolsep}{3pt}
\renewcommand{\arraystretch}{1.05}

\resizebox{\linewidth}{!}{%
\begin{tabular}{l
                S[table-format=2.3] S[table-format=1.3]
                S[table-format=2.3] S[table-format=1.3]
                S[table-format=2.3] S[table-format=1.3]
                S[table-format=2.3] S[table-format=1.3]
                S[table-format=2.3] S[table-format=1.3]
                S[table-format=2.3] S[table-format=1.3]}
\toprule
\multirow{2}{*}{Methods}
& \multicolumn{2}{c}{Chamber 1}
& \multicolumn{2}{c}{Chamber 2}
& \multicolumn{2}{c}{Chamber 3}
& \multicolumn{2}{c}{Chamber 4}
& \multicolumn{2}{c}{Chamber 5}
& \multicolumn{2}{c}{Chamber 6} \\
\cmidrule(lr){2-3}\cmidrule(lr){4-5}\cmidrule(lr){6-7}
\cmidrule(lr){8-9}\cmidrule(lr){10-11}\cmidrule(lr){12-13}
& {RMSE} & {MAE}
& {RMSE} & {MAE}
& {RMSE} & {MAE}
& {RMSE} & {MAE}
& {RMSE} & {MAE}
& {RMSE} & {MAE} \\
\midrule

LSTM~\cite{hamrani2020machine} (2020)
& 3.295 & 2.144 & 2.943 & 1.506 & 5.013 & 2.147 & 12.568 & 5.234 & 4.008 & 2.309 & 2.269 & 1.184 \\

RF~\cite{saha2021machine,liu2024co} (2021/2024)
& 3.803 & 2.151 & 3.879 & 1.969 & 5.166 & 2.266 & 8.091 & \second{2.777} & 3.937 & 2.270 & 3.913 & 1.597 \\

GRU~\cite{liu2022kgml} (2022)
& 3.622 & 2.542 & 2.954 & 1.547 & 4.903 & 2.166 & 11.159 & 4.728 & 3.543 & 2.204 & 1.990 & 0.981 \\

XGBoost~\cite{wang2024sources} (2024)
& 3.868 & 2.617 & 3.372 & 1.726 & 5.303 & 2.419 & 8.418 & 2.989 & 4.464 & 2.502 & 3.944 & 1.573 \\

MLP~\cite{vemuri2024developing} (2024)
& 5.080 & 3.683 & 5.901 & 3.507 & 7.959 & 4.293 & 12.264 & 6.620 & 6.944 & 4.143 & 4.225 & 2.895 \\

Transformer~\cite{lin2024agriculture} (2024)
& 3.209 & 2.248 & 2.835 & 1.585 & 4.483 & 2.110 & 10.288 & 4.211 & 3.481 & 2.057 & 1.806 & 0.889 \\

\midrule
DA-LSTM
& \second{2.596} & \second{1.316} & 2.849 & 1.522 & 4.749 & 1.946 & 9.932 & 4.214 & 4.010 & 2.317 & 2.165 & 1.044 \\

DA-RF
& 3.805 & 2.032 & 4.410 & 2.207 & 4.990 & 2.062 & 8.366 & 2.912 & 4.895 & 2.762 & 4.948 & 2.010 \\

DA-GRU
& 3.499 & 1.958 & 2.675 & 1.564 & 4.413 & \best{1.831} & 11.147 & 4.443 & 3.442 & 1.886 & 2.170 & 1.096 \\

DA-XGBoost
& 4.053 & 2.739 & 3.622 & 1.945 & 5.550 & 2.696 & 8.555 & 3.209 & 4.923 & 2.756 & 3.987 & 1.734 \\

DA-MLP
& 3.542 & 2.155 & 4.220 & 2.193 & 5.529 & 2.637 & 10.212 & 5.058 & 4.818 & 3.275 & 3.159 & 1.807 \\

DA-Transformer
& \best{2.262} & \best{1.055}
& \second{2.559} & \second{1.463}
& \best{3.674} & 1.928
& \second{7.396} & 4.620
& \second{3.067} & \second{1.858}
& \second{1.623} & \second{0.827} \\

\midrule
DA-Mamba (Our)
& \second{2.596} & 1.479
& \best{1.568} & \best{0.883}
& \second{4.232} & \second{1.911}
& \best{5.655} & \best{2.033}
& \best{2.779} & \best{1.735}
& \best{1.283} & \best{0.676} \\
\bottomrule
\end{tabular}%
}
\end{table}

It can be observed that, after incorporating the DA module, deep learning models (MLP, GRU, LSTM, Transformer, and Mamba) exhibit substantial performance gains.
These improvements can be attributed to the models’ enhanced capacity to capture complex temporal dependencies and latent data patterns, enabled by their nonlinear approximation capabilities and the effective utilisation of the augmented and diversified training data.
As a result, these models are better equipped to learn intrinsic features, model temporal dynamics, and capture inter-variable dependencies, thereby significantly enhancing their generalisation performance.
Notably, our proposed Mamba-based model consistently achieves state-of-the-art performance both with and without the DA module, highlighting its strong generalisation capability and robustness across diverse agricultural scenarios.
In contrast, the performance of traditional tree-based models (RF and XGBoost) deteriorates after applying DA.
The RF model, in particular, shows a noticeable drop in R$^2$.
This decline can be attributed to the fact that tree-based models depend heavily on the original data distribution to establish optimal decision boundaries.
The noise and redundancy introduced by data augmentation may disrupt the dataset’s inherent structure, thus impairing the model’s ability to identify meaningful feature relationships.

\subsection{Parameter Sensitivity Analysis for Data Augmentation}
In order to further explore the potential of the data augmentation module, we conducted a comprehensive parameter sensitivity experiment.
Specifically, the parameter $h$ is used to randomly select valid hourly observations from the 24 hours a day.
These observations are then averaged and multiplied by 24 to form a representative daily measurement. To preserve intrinsic diurnal patterns while introducing minimal changes, $h$ is chosen to be greater than 12 but less than 24.
Therefore, we set $h$ to 12, 14, 16, 18, 20, 22, and 24 as random sampling values and performed experiments using our proposed DA-Mamba.
Note that $h=24$ is equivalent to using no data augmentation.
In addition, when $h=22$, we randomly selected $n=1000$ to extend the training dataset, resulting in a total of $n \times 5 \times 3 \times 122$ days of data to ensure fairness in the comparison experiment. However, since there are only $\binom{24}{22} = 276$ unique ways to choose 22 hours out of 24, some repeated samples inevitably appear in the augmented training set.

The results of the comparison experiment are shown in Fig.~\ref{k-fold_result}(b), from which three notable trends can be observed.
Firstly, the impact of the data augmentation module on the downstream prediction task increases and decreases as the random sampling size $h$ grows from 12 to 20, forming a U-shaped trend and peaking at $h=16$ for all three metrics: RMSE, MAE, and R$^2$.
Secondly, when $h$ is increased to 22, the model achieves the second-best performance.
This indicates that although fewer unique augmented samples are generated as $h$ approaches 24 hours, these samples still contribute meaningfully to improving the accuracy of N$_2$O emission predictions.
Finally, when $h = 12$, the low sampling frequency limits the effectiveness of the data augmentation module, resulting in degraded downstream performance.

\begin{figure}[pos=!b]
\centering
\includegraphics[width=0.6\linewidth]{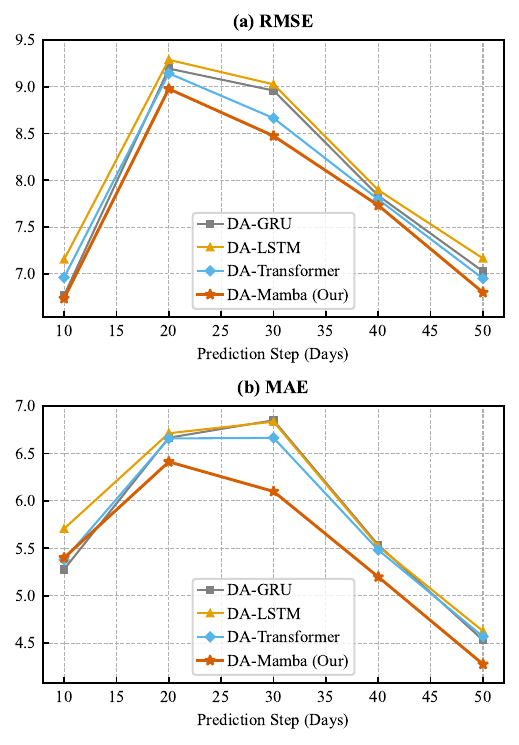}
\caption{(a) RMSE values for three baseline models with data enhancement, compared against the proposed DA-Mamba model across different prediction horizons (10, 20, 30, 40, and 50 days).
(b) MAE values for the same set of models and prediction steps.}
\label{mse_and_mae_multistep}
\end{figure}

\subsection{Multi-Step Prediction}
\begin{figure}[pos=!htbp]
\centering
\includegraphics[width=1.0\linewidth]{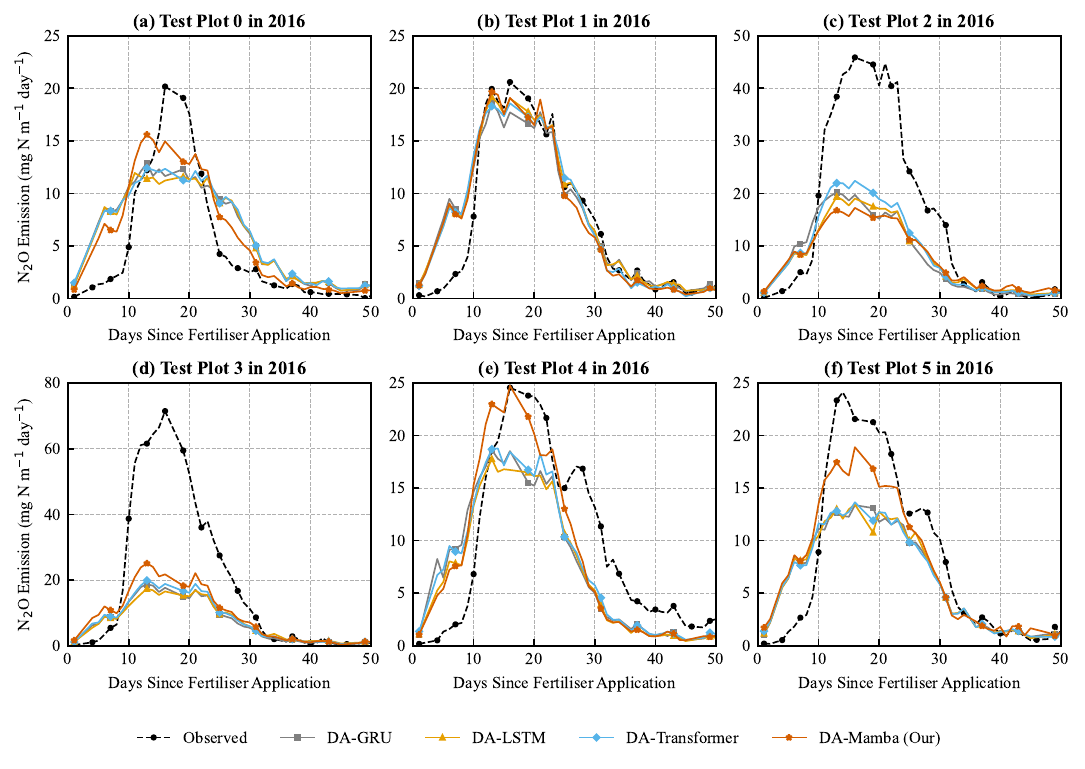}
\caption{Multi-step N$_2$O prediction results comparison on six test plots 
(0–5) in 2016 following fertiliser application. Black dashed lines denote observed values, while coloured curves correspond to predictions from different models.}
\label{MultiStepPrediction}
\end{figure}

To evaluate the performance of our proposed DA-Mamba model on the multi-step prediction task, we conducted a comparative experiment against DA-GRU, DA-LSTM, and DA-Transformer, using $h = 16$ for the data augmentation module.
As illustrated in Fig.~\ref{mse_and_mae_multistep}(a), our proposed DA-Mamba consistently achieve the lowest RMSE across all prediction steps, with its advantage becoming more pronounced at longer horizons.
The RMSE increases from prediction step 10 to 20, forming a peak, and then gradually decreases as the prediction step extends to 30, 40, and 50.
\textbf{The inverted U-shaped trend can be explained by the emission dynamics shown in Fig.~\ref{MultiStepPrediction}} following fertiliser application, N$_2$O emissions rise sharply around day 10, surging from approximately 3 to over 20 mg N m$^{-1}$ day$^{-1}$ by day 20, thus increasing the difficulty of prediction within that range.
Subsequently, emissions decline rapidly and stabilise near zero, making longer-horizon predictions more stable and easier to estimate.
As illustrated in Fig.~\ref{mse_and_mae_multistep}(b),  at the shortest prediction horizon (10 days), DA-GRU achieves the lowest MAE.
However, DA-Mamba attains the lowest RMSE, suggesting a stronger capability to capture outlier N$_2$O emission values. This highlights that DA-Mamba not only maintains overall accuracy but also handles extreme variations more effectively. As the prediction step increases, the performance gap between DA-Mamba and the baseline models (DA-GRU, DA-LSTM, DA-Transformer) progressively widens, with DA-Mamba achieving the best results in both RMSE and MAE from step 20 onwards. 
In Test Plot 0 (Fig.~\ref{MultiStepPrediction}(a)), DA-Mamba provides a smoother 
fit to the rapid rise around day 12 and captures the subsequent decline with 
lower deviation. 
In addition, Fig.~\ref{MultiStepPrediction}(b) shows that, while the predictions from all models do not differ significantly from 10 to 50 days after fertilisation, DA-Mamba more accurately captures the first peak in greenhouse gas emissions, which occurs around day 13 after fertilisation.
Fig.~\ref{MultiStepPrediction}(d) presents the N$_2$O emissions from Test Plot 3, 
which exhibit substantial variability with peaks exceeding 70 mg N m$^{-1}$ day$^{-1}$. 
As a result, none of the models is able to capture the peak accurately, though 
DA-Mamba still approximates the magnitude and timing of the surge more closely, 
showing greater temporal sensitivity and slightly smaller bias than the baseline models.
Furthermore, in Test Plot 4 (Fig.~\ref{MultiStepPrediction}(e)), DA-Mamba aligns more closely with the observed peak near day 17 and maintains lower deviation throughout the post-peak decline period, highlighting its robustness in modelling emission decay phases.
Similarly, for Test Plot 5 (Fig.~\ref{MultiStepPrediction}(f)), DA-Mamba produces a peak estimate of approximately 18 mg N m$^{-1}$ day$^{-1}$ on day 16, which is closer to the ground truth value of around 24, while all baseline models underestimate the peak at roughly 11–13 mg N m$^{-1}$ day$^{-1}$.
In summary, although DA-Mamba shows a slight disadvantage in Test Plot 2, the overall results confirm its superior generalisation and temporal sensitivity across different experimental plots, particularly in capturing both the magnitude and dynamics of sharp emission surges and subsequent decay phases. This demonstrates its suitability for real-world agricultural emission prediction and decision support.

\section{System Integration}
\begin{figure}[pos=!t]
\centering
\includegraphics[width=0.6\linewidth]{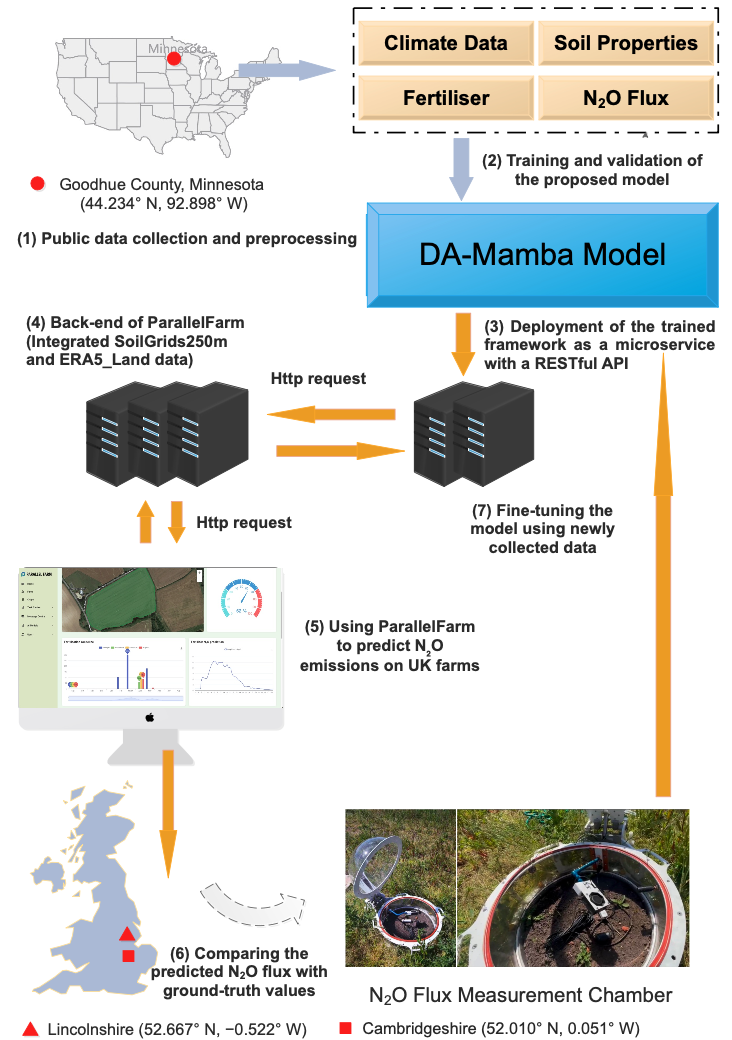}
\caption{System integration}
\label{SystemFramework}
\end{figure}

The end-to-end operation of DA-Mamba within the ParallelFarm system is illustrated in Figure~\ref{SystemFramework}.
This workflow encompasses public data curation and model training (Steps 1–2), global-scale third-party data integration, containerised deployment and backend orchestration (Steps 3–4), user-facing decision support (Step 5), field validation based on chamber measurements (Step 6), and closed-loop fine-tuning using newly acquired observational data (Step 7).
Further details are provided below:

\noindent\textbf{(1) Public data collection and preprocessing.}
We integrated four types of data blocks from the public data repository at the benchmark site in Goodhue County, Minnesota—daily climate data, static soil properties, management practices (fertiliser application), and N$_2$O flux data~\cite{liu2022kgml}. Climate sequences underwent missing value checks before being aggregated and normalised at the daily scale; static field-scale soil properties were replicated daily and aligned with weather data; fertilisation events were converted into standardised doses and temporal features.

\noindent\textbf{(2) Training and validation of DA--Mamba.}
Using the step (1) dataset, DA-Mamba is trained and validated; experiments in Section~IV confirm that it achieves state-of-the-art predictive accuracy.

\noindent\textbf{(3) Deployment as a RESTful microservice.}
The trained artefact is encapsulated in a stateless FastAPI service exposing a versioned endpoint. Pydantic schemas enforce input validity and explicit missingness; responses include daily flux trajectories (mg\,N\,m$^{-2}$\,day$^{-1}$) and metadata (model version, data snapshot hash). The image is containerised and scheduled on K3S, enabling horizontal replication under load without coupling to transactional services.

\noindent\textbf{(4) Back-end integration (SoilGrids250\,m and ERA5\_Land).}
ParallelFarm's Spring Boot back end acts as an orchestrator. For each field polygon, soil covariates are retrieved or cached; ERA5\_Land daily aggregates are bilinearly sampled at the centroid (or polygon-average) and aligned to the requested horizon. Fertiliser plans from the Records/Task modules are joined to form a canonical payload, which is dispatched via HTTP to the DA--Mamba service. Long-running scenario batches are placed on a message queue; results are persisted to MongoDB with immutable provenance.

\noindent\textbf{(5) User-side prediction and decision support.}
Via the ParallelFarm web interface, farm managers are able to create precision fertilisation plans for each field and obtain predictive insights into N$_2$O emissions.

\noindent\textbf{(6) Ground-truth comparison on UK farms.}
Model outputs are benchmarked against chamber-based N$_2$O-flux measurements collected on test plots in Lincolnshire (52.667° N, –0.522° W) and Cambridgeshire (52.010° N, 0.051° W). 

\noindent\textbf{(7) Fine-tuning with newly collected data.}
Newly collected flux data are uploaded to refine DA-Mamba, allowing dynamic adaptation to changing practices and conditions. This enables a closed-loop workflow of data ingestion, decision-making, and field validation. Leveraging global soil and climate data, the system scales easily to other farms for N\textsubscript{2}O prediction, continuously improved through in-situ observations.

\section{Discussion}
\subsection{Scalability, Governance and Limitations}
The proposed DA-Mamba framework achieves containerised deployment through integration with the ParallelFarm system, ensuring modular design and horizontal scalability to deliver real-time decision support in practical farm management scenarios. Simultaneously, by leveraging globally accessible datasets such as ERA5-Land and SoilGrids250m, the framework demonstrates strong regional scalability potential, enabling expansion from local experimental fields to broader agricultural regions and even global scales.

At the governance level, integrating DA-Mamba into agricultural management systems raises critical issues regarding data ownership, model transparency, and accountability mechanisms. As the framework relies on management records provided by farms and farmers, along with site-specific soil data, establishing a secure data governance framework is imperative to ensure privacy protection, data interoperability, and fairness in model outputs. Clear and standardised governance protocols play a pivotal role in enhancing system security, fostering trust among stakeholders, and driving large-scale adoption.

Nevertheless, the framework exhibits certain limitations. First, the limited scale and restricted geographic distribution of N$_2$O datasets from box experiments may weaken the model's generalisation ability under heterogeneous soil and climatic conditions. Although data augmentation partially mitigates overfitting issues, the framework remains reliant on limited high-quality observational data, potentially compromising robustness in regions lacking long-term monitoring. In addition, as demonstrated by multi-step prediction experiments, sharp and irregular extreme emission peaks remain challenging for all models, including DA-Mamba. Finally, despite its computational efficiency, the system faces deployment barriers in low-resource farms lacking cloud infrastructure and stable internet connectivity, particularly in developing and underdeveloped regions.

Therefore, future work should focus on three areas: Firstly, expanding field trials to enrich datasets and enhance the model's cross-regional generalisation capabilities.
Secondly, establishing standardised governance mechanisms for agricultural AI systems to improve system security and transparency.
Thirdly, exploring deployment and field data collection in developing or resource-constrained regions to support the framework's widespread implementation and application across diverse agricultural environments.

\section{Conclusion}
In this study, we proposed the DA-Mamba, which achieves state-of-the-art performance on fertiliser-induced N$_2$O emission datasets in both single- and multi-step prediction tasks.
Specifically, DA-Mamba improves single-step prediction accuracy by up to 12.0\% (RMSE) and 25.8\% (MAE), and multi-step forecasting accuracy by 2.0\% (RMSE) and 4.8\% (MAE) compared with baseline models, as validated through comprehensive empirical analyses.
By integrating DA-Mamba into the ParallelFarm system, the system is capable of automatically predicting N$_2$O emissions based on AI-generated precision fertilisation schemes.
Moreover, it establishes a closed-loop workflow that encompasses data ingestion, intelligent decision support, and field-level verification. Continuous uploads of field observations enable iterative model retraining, thereby facilitating a fully data-driven and adaptive framework for forecasting agricultural GHG emissions.
In future work, we plan to scale up the deployment of the system to a wider range of farms across the UK.
Large-scale field trials will be undertaken to gather diverse, high-quality datasets, supporting further model enhancement and validation under practical agricultural conditions.



\section*{Declaration of Competing Interest}
The authors declare that they have no known competing financial interests or personal relationships that could have appeared to influence the work reported in this paper.

\section*{Data availability}
The data used to support the findings of this study are publicly available from the Zenodo repository at \url{https://zenodo.org/records/5504533}, as described in Liu et al.~\cite{liu2022kgml}.
All experiments and evaluations reported in this paper were conducted exclusively using this publicly available dataset.
The proposed system is designed to support continuous data collection in real-world deployments; however, such newly collected data were not used in the analyses presented in this study.

\renewcommand{\bibfont}{\small}
\bibliographystyle{cas-model2-names}
\bibliography{references} 

\end{document}